\documentclass{article}

\usepackage[preprint]{neurips_2022}

\usepackage[numbers]{natbib}

\usepackage[utf8]{inputenc} 
\usepackage[T1]{fontenc}    
\usepackage{hyperref}       
\usepackage{url}            
\usepackage{booktabs}       
\usepackage{amsfonts}       
\usepackage{nicefrac}       
\usepackage{microtype}      
\usepackage{xcolor}         

\usepackage{graphicx}
\usepackage{enumerate}
\usepackage{algorithm}
\usepackage{algorithmic}
\usepackage{subfigure}
\usepackage{multirow}
\floatname{algorithm}{Algorithm}  
\usepackage{amsmath,amssymb,amsthm}

\DeclareMathOperator*{\argmax}{argmax}

\renewcommand{\P}{\mathcal{P}}

\newcommand{\R}{\mathcal{R}}

\title{Multi-Agent Reinforcement Learning with \\a Hierarchy of Reward Machines}

\author{%
  Xuejing Zheng\\
  Department of Computer Science, Sun Yat-sen University\\
  \texttt{zhengxj28@mail2.sysu.edu.cn} \\ 
  \And
  Chao Yu\\
  Department of Computer Science, Sun Yat-sen University\\
  \texttt{yuchao3@mail.sysu.edu.cn}
}

\begin{document}
\maketitle

\begin{abstract}
    In this paper, we study the cooperative \textit{Multi-Agent Reinforcement Learning} (MARL) problems using \textit{Reward Machines} (RMs) to specify the reward functions such that the prior knowledge of high-level events in a task can be leveraged to facilitate the learning efficiency. Unlike the existing work that RMs have been incorporated into MARL for task decomposition and policy learning in relatively simple domains or with an assumption of independencies among the agents, we present \textit{Multi-Agent Reinforcement Learning with a Hierarchy of RMs} (MAHRM) that is capable of dealing with more complex scenarios when the events among agents can occur concurrently and the agents are highly interdependent. 
    MAHRM exploits the relationship of high-level events to decompose a task into a hierarchy of simpler subtasks that are assigned to a small group of agents, so as to reduce the overall computational complexity. 
    Experimental results in three cooperative MARL domains show that MAHRM outperforms other MARL methods using the same prior knowledge of high-level events.
\end{abstract}

\section{Introduction}

Cooperative \textit{Multi-Agent Reinforcement Learning} (MARL)~\cite{gronauer2022multi} studies the learning problems when multiple agents interact with each other to complete a common task in an unknown environment. 
Since multiple agents learn and update their policies concurrently in the same  environment, it is difficult to learn a satisfactory policy due to the arising issue of non-stationary~\cite{hernandez2017survey,papoudakis2019dealing}.
Previous works on MARL mitigate this issue by learning centralized value functions, but still suffer from the problems when rewards are either sparse or cannot be modeled as Markovian ones. 
As a type of finite state automata, \textit{Reward Machines} (RMs)~\cite{icarte2018using} have been incorporated into cooperative MARL in order to exploit task decomposition and high-level ideas or prior knowledge to overcome these issues.
\citet{neary2021reward} used a predefined team RM to specify agents' common task, and proposed \textit{projections of RM} so as to decompose the team RM into sub-RMs representing the local subtasks for the agents. \citet{hu2021decentralized} used a graph to represent neighborly relationship of agents, and studied the graph-based MARL where the transition of each agent only depends on its neighboring agents. 
The local reward function of each agent is then specified by an RM to cooperatively complete the common task.
In these works, the joint state and action space are split into local ones forcibly with the aid of decomposing the team task through RMs, thus avoiding the dimensional curse issue in general MARL problems.
Besides, high-level events or prior knowledge are shared and synchronized among the agents during training and execution, which enables effective coordination among the agents and thus alleviates the non-stationary issue. 

However, current works utilizing RMs to solve the cooperative MARL problems still suffer from several limitations. 
First, since the size of local RM of each agent should be relatively small to ensure efficient learning, these works have to rely on the assumption of weak interdependencies among the agents.
In scenarios when the agents are highly interdependent, the size of local RM of each agent can be close to the size of the team RM, which grows dramatically with the number of agents. 
Second, such methods are limited to scenarios involving relatively simple prior knowledge, which means that each high-level event of RM only consists of a single proposition such as \textit{``agent 1 presses button''} and \textit{``agent 2 presses button''}, rather than multiple propositions like \textit{``both agent 1 and 2 press button''}. Such limitation in exploiting more complex prior knowledge makes these methods difficult to specify tasks with multiple concurrent events. 

In this paper, we propose \textit{Multi-Agent reinforcement learning with Hierarchy of RMs} (MAHRM) for more efficient cooperative MARL by utilizing the prior knowledge of high-level events and specifying the rewards of a task as a hierarchy of RMs. 
In order to reduce the size of RM of a task when the agents are highly interdependent, a hierarchical structure of propositions (\textit{i.e.} high-level events) is leveraged to decompose a complex task into simple subtasks that are assigned to a small group of agents, so as to reduce the size of the states and action spaces in lower-level subtasks.
Each proposition in this structure represents a subtask, and a higher-level proposition is a temporal abstraction of lower-level propositions.
Then the MAHRM is trained in a way similar to \textit{Hierarchical Reinforcement Learning} (HRL): the policy of a subtask represented by the proposition at higher-level chooses lower-level subtasks to be executed, and the lowest-level policies determine the actions for all the agents to be executed in the environment. Experimental results in three domains show that MAHRM outperforms baselines that utilize the same prior knowledge of high-level events, including the RM and HRL approaches for cooperative MARL.

\section{Preliminaries}
\label{sec:preliminaries}
\textbf{Cooperative MARL}~solves the problem of how multiple agents simultaneously interact with each other to complete a common task in an unknown environment. The underlying model for cooperative MARL is \textit{fully cooperative Markov Games}, formalized by a tuple $M=(N,S,A,P,R,\gamma)$, where $N$ is the number of agents; $S=S_1\times \cdots \times S_N$ is a joint state space, and $S_i$ denotes the set of states of agent $i$; $A=A_1\times\cdots\times A_N$ is a joint action space, and $A_i$ denotes the set of actions of agent $i$; $P:S\times A\to \Delta(S)$ is a probabilistic transition function, and $\Delta$ denotes the probability distribution; $R:S\times A\times S\to \mathbb{R}$ is a shared reward function and $\gamma\in [0,1]$ is a discount factor. We use $s$ and $a$ to denote the vector of states and actions for all agents, respectively. The local state and action of agent $i$ is denoted as $s_i$ and $a_i$, respectively. A policy of cooperative MARL is a mapping $\pi:S\to \Delta(A)$, and $\pi(a\mid s)$ means the probability of choosing action $a$ for all agents in state $s$. 

\textbf{Q-learning} is a well known approach to calculate the optimal policy via Q-value iterations in tabular case~\cite{watkins1992q}. The Q-function $Q^\pi(s,a)$ following the policy $\pi$ is the expected discounted reward of choosing action $a$ in state $s$ under policy $\pi$, formalized as $Q^\pi(s,a)=\mathbb{E}_\pi[\sum_{k=0}^\infty \gamma^k r_{t+k}\mid s_t=s,a_t=a]$, where $r_k=R(s_k,a_k,s_{k+1})$, $s_{k+1}$ and $a_k$ are sampled from probabilistic transition $P(s_k,a_k,s_{k+1})$ and policy $\pi(a_k\mid s_k)$, respectively. The one-step updating rule of Q-learning is given by $Q(s,a)\xleftarrow{\alpha}r(s,a,s')+\gamma \max_{a'} Q(s',a')$, where $x\xleftarrow{\alpha}y$ means $x\leftarrow x+\alpha(y-x)$. The action $a$ is chosen by using certain exploration strategies, such as the $\epsilon$-greedy policy, \textit{i.e.} choosing a random action with probability $\epsilon$, while choosing $\argmax_{a'}Q(s,a')$ with probability $1-\epsilon$.  

\textbf{Multi-Agent Hierarchical Reinforcement Learning (MAHRL)}~\cite{ghavamzadeh2006hierarchical} decomposes a long-horizon MARL task into several subgoals, skills or subtasks, and maintains a high-level policy to choose optimal subtasks to perform at lower levels, while the low-level policies of agents learn to solve these subtasks explicitly~\cite{pateria2021hierarchical}. The MODULAR framework~\cite{lee2019learning} is a cooperative MAHRL approach, which is composed of two components: a meta policy $\pi_{meta}$ and $N$ subtasks of $N$ agents. The meta policy learns to select a subtask for each agent to execute, and each agent learns its own primitive subtasks independently. 

\textbf{RMs} can be used to reveal the structure of non-Markovian reward function of a task when given the prior knowledge of high-level events. The prior knowledge consists of a set of \textit{propositions} or atomic events $\P$, a set of \textit{labels} or concurrent events $\Sigma\subseteq 2^\P$ which may possibly occur during the process of the task and the \textit{labelling function} $L:S\times A\times S\to \Sigma$ giving the abstraction of one-step experience. An RM is then formalized as a tuple $\R=(\P,\Sigma,U,u_0,F,\delta,R)$, where $U$ is a finite set of states, $u_0$ is an initial state, $F\subseteq U$ is a set of terminal states, $\delta: U\times \Sigma\to U$ is a transition function, and $R:U\times\Sigma\to\mathbb{R}$ is a reward function. The \textit{Q-learning for RM} (QRM) algorithm~\cite{icarte2018using} maintains a Q-function $Q^u$ for each RM state $u\in U$, and updates all the Q-functions using \textit{internal} rewards and transitions of RM with one experience $(s,a,s')$. Its one-step updating rule is given by
\begin{equation}
    Q^u(s,a)\xleftarrow{\alpha} R(u,l)+\gamma \max_{a'}Q^{u'}(s',a'), \text{for all $u\in U$},
\end{equation}
where $l=L(s,a,s')\in \Sigma$ is the current label given by the predefined labelling function; $R(u,l)$ is the internal reward and $u'=\delta(u,l)$ is the internal next state of RM. QRM not only converges to an optimal policy in tabular cases, but also outperforms the \textit{Hierarchical Reinforcement Learning} (HRL) methods which might converge to suboptimal policies~\cite{icarte2018using}.

\section{The MAHRM Framework}
In this section, we introduce our MAHRM framework. MAHRM exploits the relationship between propositions (\textit{i.e.} high-level atomic events) for a more compact and efficient representation of task decomposition by organizing a hierarchical structure, where a higher-level proposition is a temporal abstraction of lower-level propositions.
Each proposition represents a subtask, which is assigned to a group of agents that coordinate to make the proposition become true. In particular, the highest level proposition represents the joint task assigned to all the agents.
Then, the hierarchy of RMs is derived from the hierarchical structure of propositions by using an RM to define the reward function of subtask for each proposition.
MAHRM then learns its policies in a way similar to HRL: the policy of each higher-level subtask chooses the subtasks of propositions at lower levels to be executed, and the policies at the lowest level determine the actions for all the agents to interact with the environment. The internal structure of each RM is then exploited to eliminate unreasonable selection of subtasks, in order to  improve the efficiency of the learning process. Moreover, MAHRM enables dynamic decomposition of the joint task by assigning different subtasks to the agents at different timestep, so that the state and action spaces are split into the local ones for the interdependent agents. Hence, the agents are coordinated by higher-level policies in MAHRM to complete the joint task efficiently.

\subsection{The Hierarchy of RMs}
\label{subsec:HRM}
This subsection gives a detailed description of the hierarchy of RMs. We first introduce the hierarchical structure of propositions, which is represented as a directed acyclic graph with $K$ levels ($K\geq 2$). Each node in the graph is a proposition, and each edge connects a higher-level proposition with a relevant lower-level proposition. The set of propositions at level-$k$ is denoted by $\P_k$, and the lowest level (level-1) propositions $\P_1$ are called \textit{primitive propositions}, while propositions at higher levels (level-2,$\cdots$,level-$K$) are called \textit{non-primitive propositions}. The set of propositions connected by the edges from a non-primitive proposition $p\in \P_k$ is denoted as $\P_p$, which is a subset of $\P_{k-1}$. Each proposition represents a subtask, and a group of agents are dynamically assigned to cooperatively fulfill this subtask to make this proposition become true. We denote the proposition $p$ with a group of assigned agents $ag$ as $p(ag)$, which means that the agents $ag$ are assigned to complete the subtask of~$p$ currently. 


\begin{figure}[ht]
    \centering
    \subfigure[]{\label{fig:PassRoom}
    \includegraphics[width=0.5\linewidth]{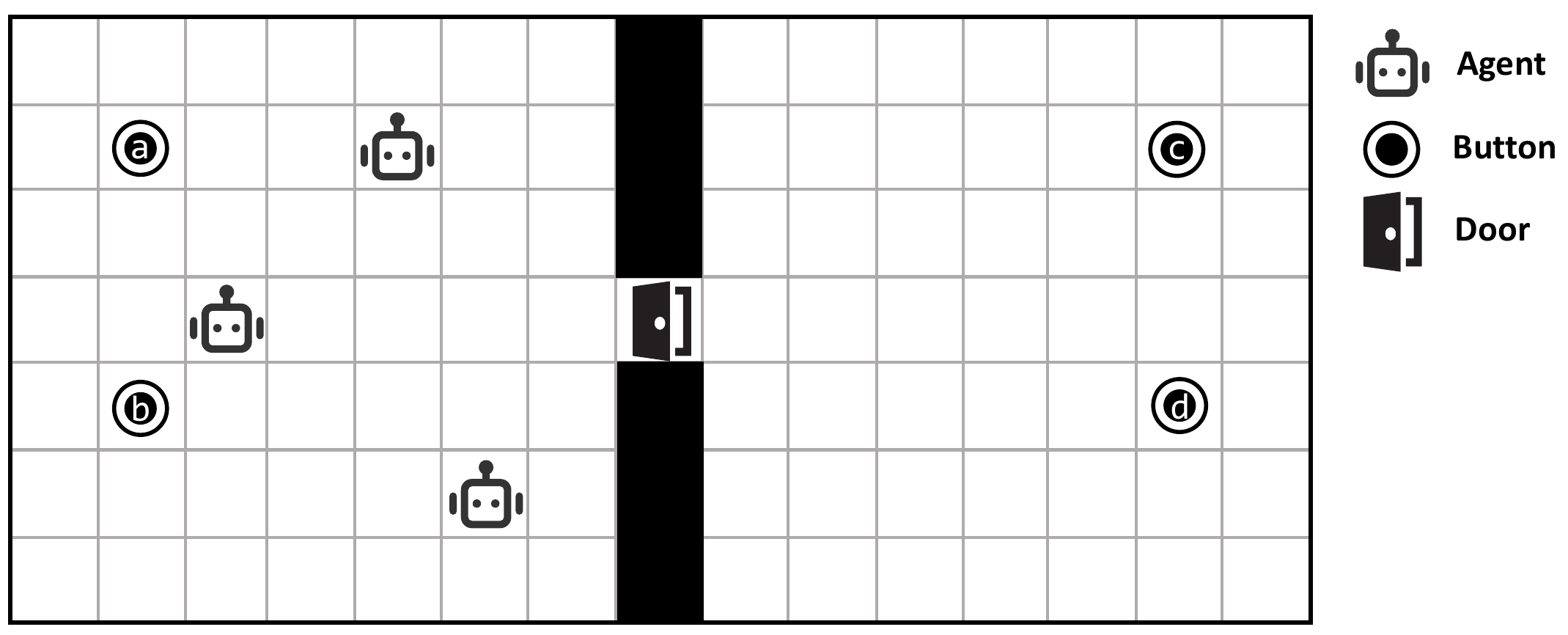}}
    \subfigure[]{\label{fig:hie_propositions}
    \includegraphics[width=0.4\linewidth]{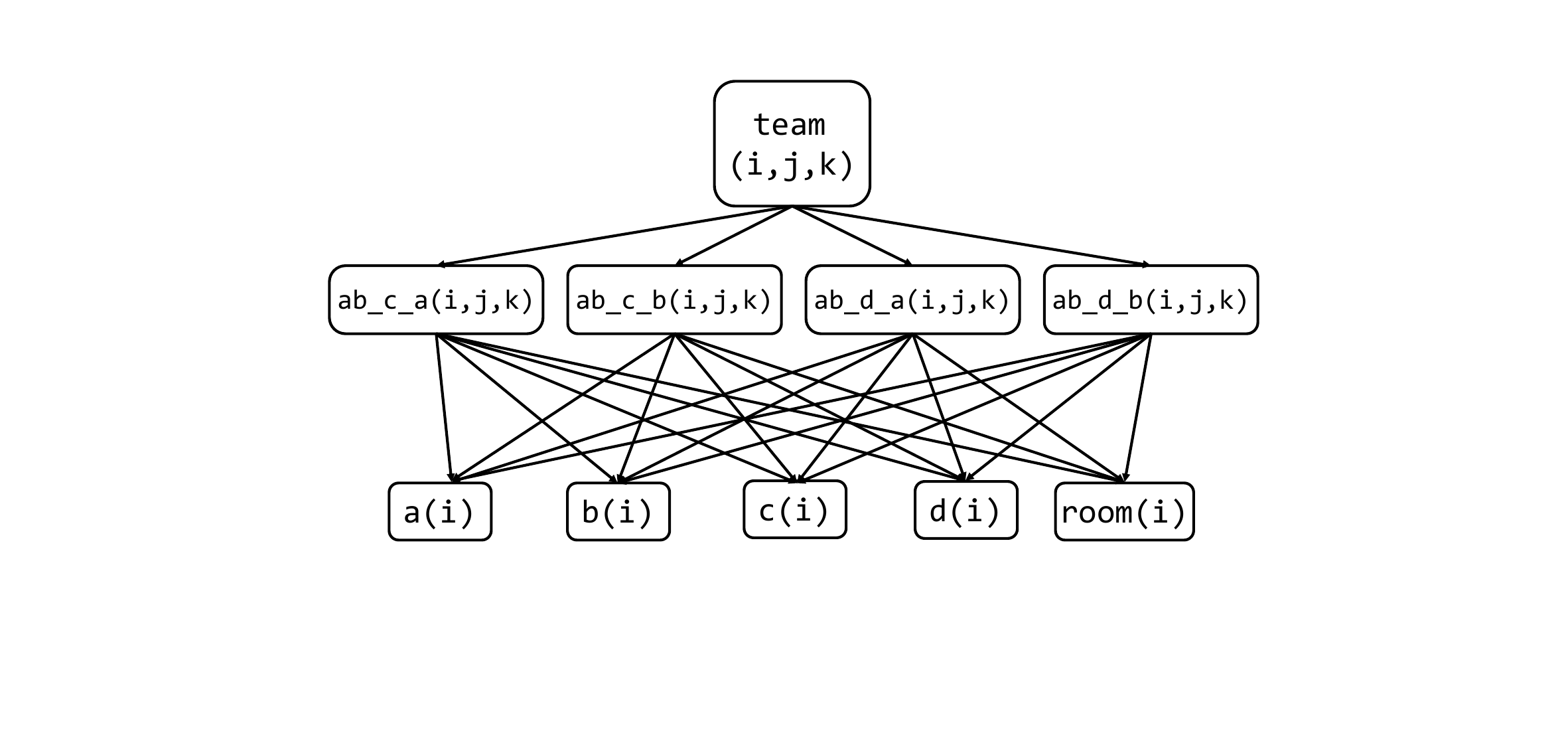}}
    \caption{(a) The \textsc{Pass} domain with three agents and four buttons tagged by \texttt{a,b,c} and \texttt{d}. (b) The 3-level hierarchical structure of propositions in the \textsc{Pass} domain.}
\end{figure}

The \textsc{Pass} domain in Figure~\ref{fig:PassRoom}~\cite{wang2019influence} is then introduced as an illustrative example, where the joint task is for three agents to pass the door and reach another room. The door is open if and only if two buttons of the four tagged \texttt{a,b,c,d} are pressed. Figure \ref{fig:hie_propositions} gives $K=3$ levels of hierarchical structure of propositions in this domain. The propositions at each level are $\P_1=\{\texttt{a(i),b(i),c(i),d(i),room(i)}\},\P_2=\{\texttt{ab\_c\_a(i,j,k),ab\_c\_b(i,j,k),ab\_d\_a(i,j,k),ab\_d\_b(i,j,k)}\}$ and~$\P_3=\{\texttt{team(i,j,k)}\}$, where \texttt{i,j,k} are agents' IDs. Each primitive proposition of \texttt{a(i),b(i),c(i)} and \texttt{d(i)} becomes true if and only if the button of \texttt{a,b,c} and \texttt{d} is pressed by agent \texttt{i}, respectively. Besides, proposition \texttt{room(i)} becomes true if and only if agent \texttt{i} passes to another room. Proposition \texttt{ab\_c\_a(i,j,k)} represents a solution to let three agents reach the room: \textit{``agent i and j press buttons a and b, respectively, to let agent k reach the room; then keeping button a pressed by agent i, agent k press button c; and finally agent j presses button d to let agent i reach the room''}. Other propositions at level-2 are defined analogously, indicating other solutions to complete the joint task. At last, the highest level proposition \texttt{team(i,j,k)} represents the joint task, which becomes true if one of the propositions at level-2 becomes true.

\begin{figure}[ht]
    \centering
    \subfigure[]{\label{fig:primitive_rm}
    \includegraphics[width=0.25\linewidth]{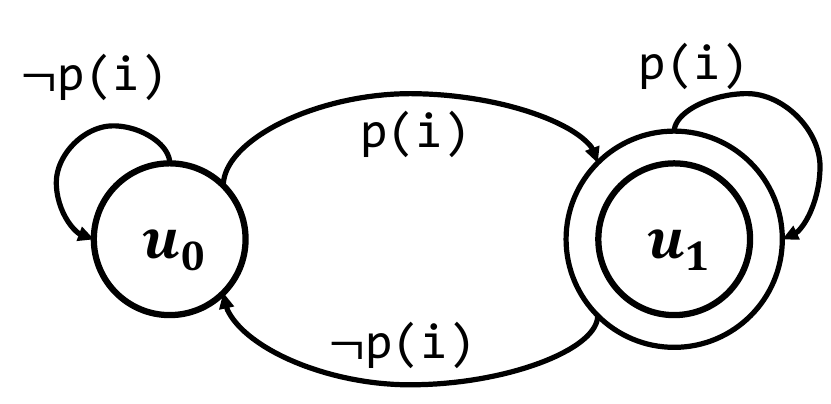}}
    \subfigure[]{\label{fig:ab_c_a}
    \includegraphics[width=0.45\linewidth]{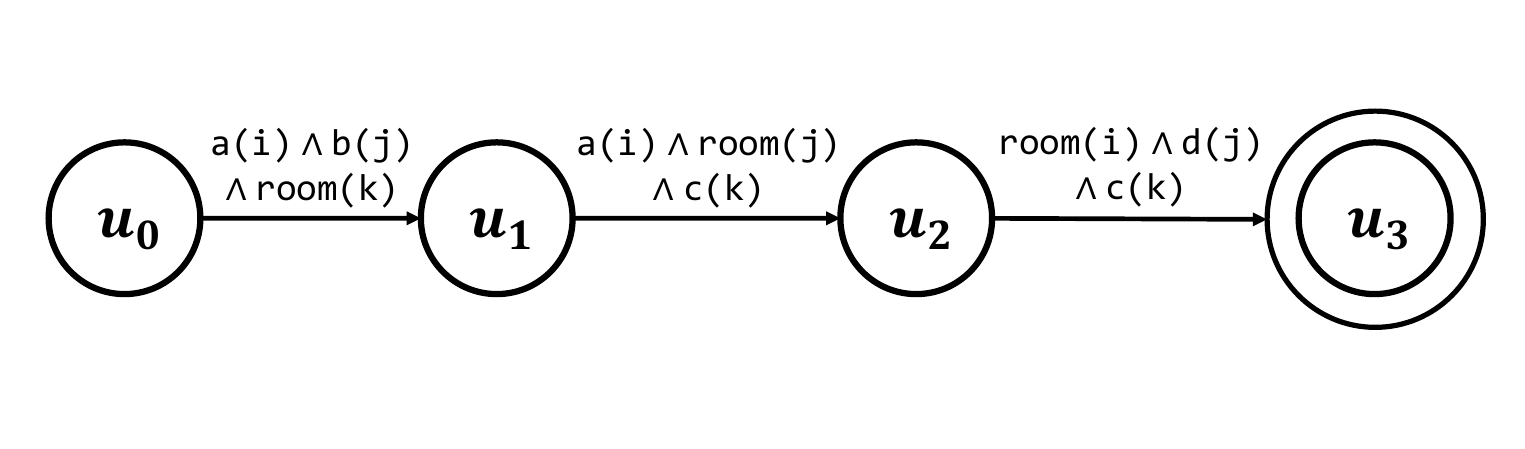}}
    \subfigure[]{\label{fig:team_rm}
    \includegraphics[width=0.25\linewidth]{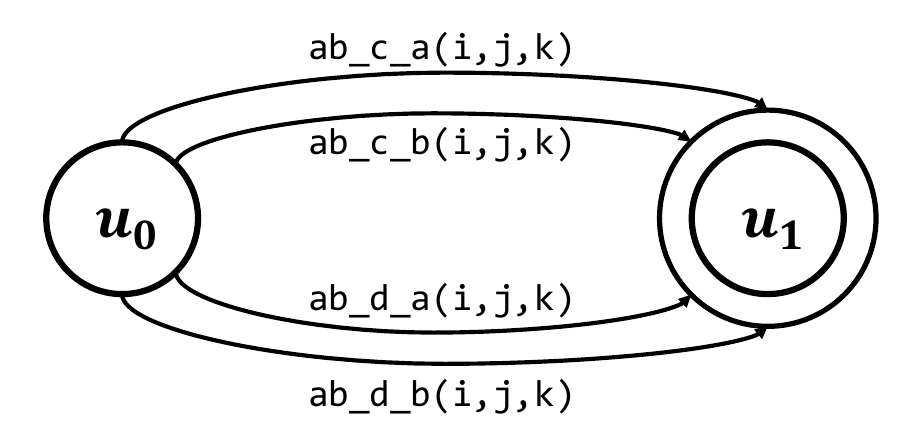}}
    \caption{(a) For a primitive proposition \texttt{p(i)}, its RM is denoted as $\R_{\texttt{p}}[\texttt{i}]$ and is constructed as follows: the states are $U=\{u_0,u_1\}$, and it transits to the terminal state $u_1$ if and only if the proposition \texttt{p(i)} becomes true, otherwise transits to $u_0$. (b) The RM of the subtask of proposition \texttt{ab\_c\_a} assigned to agents \texttt{i,j,k}. At the beginning, if the agent \texttt{i} and \texttt{j} press the button \texttt{a} and \texttt{b} so that the agent \texttt{k} reaches the room, then the RM transits to $u_1$. Then keeping the button \texttt{a} pressed, the agent \texttt{k} presses the button \texttt{c} to enable agent \texttt{j} to reach the room and the RM transits to $u_2$. Finally, the agent \texttt{j} presses the button \texttt{d} to let agent \texttt{i} reaches the room and the RM terminates at $u_3$. (c) The RM of the joint task consists of the initial state $u_0$ and the terminal state $u_1$, and its transitions are specified by propositions at level-2. The joint task is completed if one of the subtask of proposition \texttt{ab\_c\_a,ab\_c\_b,ab\_d\_a} or \texttt{ab\_d\_b} is completed.}
\end{figure}

A hierarchy of RMs is then derived from a hierarchical structure of propositions by using an RM to define the rewards of the subtask for each proposition $p$, denoted by $\R_p$. If the subtask of $p$ is assigned to agents $ag$, then the RM is denoted as $\R_p[ag]$. The set of states $U_p$ and transition function $\delta_p:U_p\times 2^{\P_p}\to U_p$ indicate the process of completing the subtask, and the reward function is defined as: $R_p(u_p,l)=1$ if and only if $u_p\not\in F_p,\delta_p(u_p,l)\in F_p$ where $F_p$ is the set of terminal states of $\R_p$, and $R_p(u_p,l)=0$ otherwise. Especially, for each primitive proposition $p\in \P_1$, the corresponding RM $\R_p$ can be automatically constructed as shown in Figure~\ref{fig:primitive_rm}, without the need of handcrafted states and transitions. As illustrative examples, the RM of non-primitive proposition \texttt{ab\_c\_a(i,j,k)} in the \textsc{Pass} domain is illustrated in Figure \ref{fig:ab_c_a}, and the RM of the joint task is shown in Figure \ref{fig:team_rm}.

\subsection{The Definition of Policies in MAHRM} 
In MAHRM, a policy of a subtask represented by the non-primitive proposition at higher levels is to select a combination of subtasks at lower levels, and then assign these subtasks to the relevant agents. The selections and assignments of subtasks are formalized as \textit{options}. The hierarchy of RMs is then used to define the set of available options of each subtask at different levels, and some unreasonable options that cannot progress the subtasks can be removed from the whole option space for more efficient learning. Besides, a policy of subtask represented by a primitive proposition at the lowest level is to choose a local action for an agent to be executed in the environment. 

Formally, for a non-primitive proposition $p\in \P_k$ at level-$k$ ($k\geq 2$), the policy of its subtask is defined as $\pi_p:U_p\to \Delta(\Omega_p)$, where $U_p$ are the states of the RM $\R_p$ and $\Omega_p$ is the \textit{option space}, \textit{i.e.} the set of all possible selections and assignments of subtasks. The RM of a subtask represented by proposition $p$ that is assigned to agents $ag_p$ is denoted as $\R_p[ag_p]$, then an option of the policy $\pi_p$ can be expressed by $o=(\R_{q_1}[ag_{q_1}],\cdots,\R_{q_m}[ag_{q_m}])$, where $m$ is the number of chosen subtasks at next level, $q_i$ is the proposition in $\P_p\subseteq \P_{k-1}$ and $ag_{q_i}$ are the agents assigned to complete the subtask of $q_i$. The option space is then eliminated by adding two constraints under each RM state $u\in U_p$: (i) the set of assigned agents to the chosen subtasks $\{ag_{q_i}\}$ is a partition of the agents $ag_p$ assigned to the subtask of $p$; and (ii) the option helps the RM $\R_p$ transit to another state to progress the subtask of $p$, which is formalized as follow:
\begin{equation}
    \Omega_{p}(u)=\left\{(\R_{q_1}[ag_{q_1}],\cdots,\R_{q_m}[ag_{q_m}])\mid q_i\in \P_{p},\{ ag_{q_i}\} \text{is a partition of }ag_p, \delta_p(u,\cup_{i=1}^m q_i)\neq u \right\},
\end{equation}
where $\delta_p: U_p\times 2^{\P_p}\to U_p$ is the transition function of $\R_p$.
Finally, for a primitive proposition $p\in \P_1$, the policy of its subtask is defined as $\pi_p:U_p\times S_i\to \Delta(A_i)$, where $i$ is the associated agent of $p$, and $S_i$, $A_i$ are its local state space and action space, respectively. 

\subsection{Learning the Policies in MAHRM}
\begin{algorithm}[t]
\caption{\texttt{MAHRM\_AT\_LEVEL}($k, o_k$)}
\label{alg:MAHRM_level}

\textbf{Input}: level $k$, option $o_k=(\R_{p_1}[ag_{p_1}],\cdots,\R_{p_m}[ag_{p_m}])$
\begin{algorithmic}[1] 
    \STATE initialize discounted accumulative rewards $\vec G_{k+1}\leftarrow 0$, option step $\tau_k\leftarrow 0$
    \STATE initialize the chosen RMs $\R_{p_i}$
    \WHILE{$t<$ max\_episode\_length \textbf{and not} current option $o_k$ terminates}
        \IF{$k=1$} 
            \STATE for each agent $i$, get action $a_i\sim$ $\pi_{p_i}(a_i\mid s_i,u_{p_i})$ 
            \STATE execute $a=(a_1,\cdots,a_N)$, observe $s'=(s_1',\cdots, s_N')$ and $l_0=L(s,a,s')$
            \STATE $\vec G_k\leftarrow (R_{p_1}(u_{p_1},l_0),\cdots,R_{p_N}(u_{p_N},l_0))$
            \STATE $t\leftarrow t+1, \vec \tau \leftarrow \vec \tau+1$
        \ELSE
            \STATE choose option $o_{k-1}=(o_{k-1}^1,\cdots,o_{k-1}^m)$, where $o_{k-1}^i\sim \pi_{p_i}(o_{k-1}^i\mid s,u_{p_i})$ 
            \STATE $\vec G_k, l_{k-1}\leftarrow$ \texttt{MAHRM\_AT\_LEVEL}($k-1,o_{k-1}$)
        \ENDIF
        \STATE \texttt{UPDATE\_RM\_STATE}($k, l_{k-1}$)
        \STATE \texttt{UPDATE\_Q\_FUNCTIONS}($k, \vec G_k$) (Equation \ref{eq:update_q_function})
        \STATE $l_{k}\leftarrow$\texttt{GET\_LABEL}($k$)
        \STATE $\vec G_{k+1} \leftarrow$\texttt{CALCULATE\_G}($k+1,\vec G_{k+1}, l_k$)  (Equation \ref{eq:calculate_G})
    \ENDWHILE
    \STATE \textbf{return $\vec G_{k+1}, l_{k}$}
\end{algorithmic}
\end{algorithm}

The process of training the policies of RMs is implemented in a recursive way based on their hierarchy, starting from the highest level-$K$ and ending at the lowest level-1 where the agents interact with the environment. Algorithm \ref{alg:MAHRM_level} gives the MAHRM training process at level-$k$. After initialization (Line 1-2), the training process proceeds until the episode or the current option terminates (Line 3-17). The option $o_k$ terminates if one of the two following conditions holds: (i) its option step $\tau_k$ exceeds the maximum option length; or (ii) one of the states of RMs at level-$(k+1)$ transits to another state.

At the lowest level $k=1$ (Line 4), the executed option $o_1$ is a combination of $m=N$ subtasks of primitive propositions $p_i$, where each subtask is assigned to a single agent $ag_{p_i}=i$, \textit{i.e.} $o_1=(\R_{p_1}[1],\cdots,\R_{p_N}[N])$.
Each agent gets action $a_i$ from policy $\pi_{p_i}$ under its current local environment state $s_i$ and RM state $u_{p_i}$ (Line 5). Then, all agents execute the joint action $a$ to interact with the environment, and observe the next joint state $s'$ from the environment and a concurrent event $l_k\subseteq \P_1$ given by the labelling function (Line 6). The one-step rewards of all agents $\vec G_k$ are returned by the chosen RMs (Line 7), which are used to update Q-functions at this level (Line 14). 


At a higher-level $k\geq 2$ (Line 9), the executed option $o_k$ is a combination of $m$ subtasks, where each subtask of $p_i$ is assigned to a group of ordered agents $ag_{p_i}$, \textit{i.e.} $o_k=(\R_{p_1}[ag_{p_1}],\cdots,\R_{p_m}[ag_{p_m}])$. The option at the next level $o_{k-1}$ is chosen according to policies $\pi_{p_i}$ of subtasks of $p_i$ (Line 10), and then executed by recursively implementing this algorithm (Line 11). 

Once the training process at level-$(k-1)$ is completed, the $\tau_k$-step accumulative discounted rewards of $m$ subtasks $\vec G_k=(G_k^1,\cdots,G_k^m)$ and events $l_{k-1}$ occurred at level-$(k-1)$ are returned (Line 11), which are utilized to update RM states and Q-functions at level-$k$ (Line 13). The states of RMs are updated by their internal transition functions, \textit{i.e.} $u_{p_i}\leftarrow \delta_{p_i}(u_{p_i},l_{k-1})$. 
At the lowest level, the Q-functions are updated by QRM algorithm using one-step rewards, while at a higher-level, $k\geq 2$ the Q-functions are updated by $\tau_k$-step Q-learning:
\begin{equation}
\label{eq:update_q_function}
    Q_{p_i}(u_{p_i,old},o_k^i)\xleftarrow{\alpha} G_k^i + \gamma^{\tau_k}\max_{o^i}Q_{p_i}(u_{p_i},o^i), \text{for all $i=1,\cdots,m$}
\end{equation}
where $Q_{p_i}$ is the Q-function of policy $\pi_{p_i}$, $u_{p_i,old}$ is the RM state $\tau_k$ steps before and $u_{p_i}$ is the current RM state. 

After the RMs at level-$k$ are updated, the label $l_k$ of this level is given by terminations of these RMs (Line 15), \textit{i.e.} $l_k=\{ p\in \P_k \mid \R_p \text{ reaches the terminal state}\}$. Line 16 then iteratively calculates the discounted accumulative reward $\vec G_{k+1}$ at higher-level by:
\begin{equation}
\label{eq:calculate_G}
    G_{k+1}^i \leftarrow  G_{k+1}^i + \gamma^{\tau_{k+1}} R_{q_i}(u_{q_i},l_k), \text{for all $i=1,\cdots,m'$}
\end{equation}
where $R_{q_i}$ and $u_{q_i}$ are the reward function and state of RM $\R_{q_i}$ at level-$(k+1)$, respectively, given by the option executed at last level $o_{k+1}=(\R_{q_1},\cdots,\R_{q_m'})$, and $m'$ is the number of chosen subtasks. Without loss of generality, we define $o_{K}=(\R_{\texttt{team}})$ as the input for implementing Algorithm~\ref{alg:MAHRM_level} at the highest level.

\section{Experiments}


In this section, we compare MAHRM to three baselines that leverage the same prior knowledge of high-level events. The first is \textit{Decentralized Q-learning with Projected RMs} (DQPRM)~\cite{neary2021reward}, which utilizes RM projections to decompose the RM of the joint task into a local RM for each agent.
Then these decentralized agents learn with their corresponding projected RMs by implementing the QRM algorithm. 
The second is \textit{Independent Q-learning with RMs} (IQRM), which shares the RM of the joint task with all agents, and then each agent learns its local policy $\pi_i:S_i\times U\to \Delta(A_i)$ independently by QRM, where $S_i$ denotes its local state space and $A_i$ denotes its action space, and $U$ denotes the set of states of the shared RM for the joint task. The third is the two-level modular framework (MODULAR)~\cite{lee2019learning} mentioned in Sec. \ref{sec:preliminaries}. 
To better leverage the prior knowledge, we modify this framework and define the set of reward functions of subtasks for agent $i$ as $\{R_{p_i}\mid p_i\in \P_i\}$, where $\P_i$ is the set of propositions related to agent $i$, and $R_{p_i}$ is the reward function of each subtask.
Concretely, $R_{p_i}$ is defined as: $R_{p_i}(s_i,a_i,s_i')=1$ if and only if $L(s_i,a_i,s_i')=\{p_i\}$, where $(s_i,a_i,s_i')$ is the local one-step experience of agent $i$, and $R_{p_i}(s_i,a_i,s_i')=0$ otherwise. 

Experiments are conduct in three domains: the \textsc{Navigation}~\cite{lowe2017multi}, the \textsc{MineCraft}~\cite{andreas2017modular} and the \textsc{Pass} domain~\cite{wang2019influence}. 
Detailed descriptions of these domains are demonstrated in the following subsections. The parameters in these domain are set as $\epsilon=0.1,\alpha=0.1$, and the maximum option length is set as $50$ for MAHRM and MODULAR. The discount factor is set as $\gamma=0.9$ in the \textsc{Navigation} and \textsc{MineCraft} domain, and $\gamma=0.95$ in the \textsc{Pass} domain. We report the performance by calculating the number of testing steps to complete the joint task for every 1000 training steps. The final results are the median performance and 25\%-75\% percentiles over 10 independent trails.

\subsection{The \textsc{Navigation} Domain}
\begin{figure}[t]
    \centering
    \includegraphics[width=\linewidth]{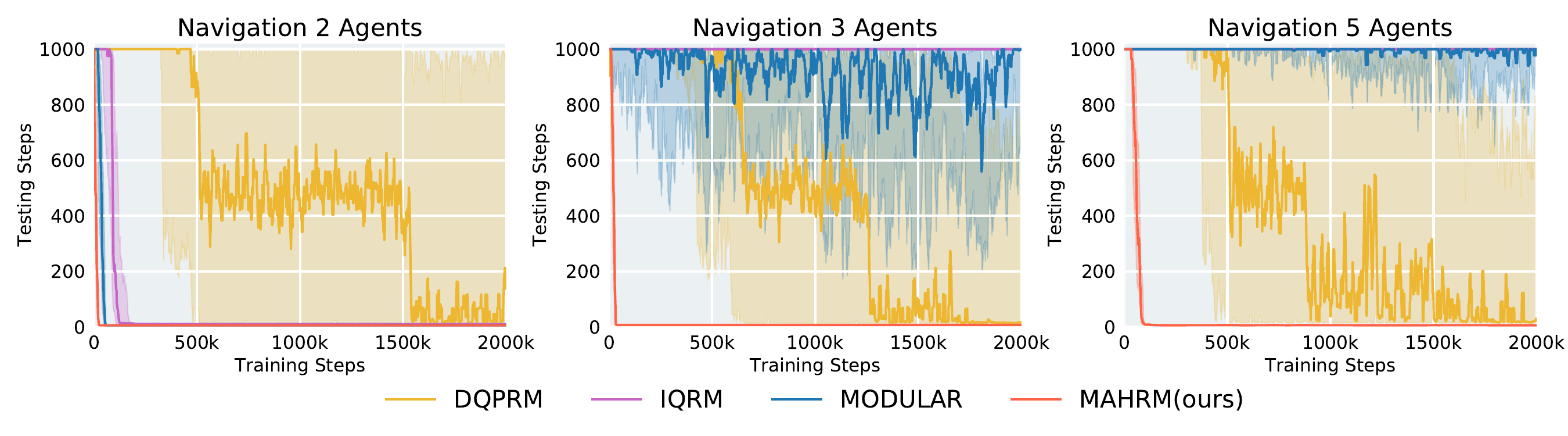}
    \caption{Experimental results in the \textsc{Navigation} domain for $N=2,3$ and 5 agents.}
    \label{fig:nav}
\end{figure}

In this domain, $N$ agents coordinate to reach a set of $N$ landmarks. The prior knowledge of this domain is given by associating each landmark with a proposition $p$, and $p(i)$ becomes true if agent $i$ reaches that landmark. We evaluate the performance of MAHRM and the baselines in the cases where $N=2,3$ and $5$, and the experimental results are shown in Figure \ref{fig:nav}. As can be seen, MAHRM outperforms other baselines in all these settings. Both IQRM and MODULAR outperform DQPRM in the setting of two agents. However, as the number of agents increases, the subtask space in MODULAR grows dramatically, while the local policies of IQRM suffer from non-stationary. Hence, these methods fail to learn good policies in the settings with a larger number of agents. DQPRM achieves good performance when the number of training steps is sufficiently large in all the cases, but the performance is unstable since the projected RMs cannot deal with the concurrent events that the agents simultaneously reach the landmarks.

\subsection{The \textsc{MineCraft} Domain}
We use the multi-agent version of \textsc{MineCraft} domain modified from \citet{leon2020extended}, where the agents collect various raw objects to manufacture new objects. 
Compared to the original version used in \cite{leon2020extended}, we further consider concurrent high-level events since the agents may collect different objects simultaneously. 
The propositions used in this domain are the tags of the raw objects, and the proposition $p(i)$ becomes true if agent $i$ gets the raw object tagged with $p$. 
To verify the effectiveness of our method in dealing with concurrent events, we define a joint task where three interdependent agents pick objects simultaneously. The RM of this task is illustrated in Appendix A.
Experimental results in Figure~\ref{fig:exp2} demonstrate that MAHRM performs the best in this domain, while IQRM suffers from non-stationarity as the number of agents increases, and DQPRM fails to decompose the joint task effectively due to the concurrent events. As for MODULAR, since its meta policy requires the rewards of the joint task to be Markovian, it is unable to learn to choose the correct subtasks under the current joint state.
MAHRM exploits the structure of joint task RM and eliminates unreasonable selections of subtasks, and thus can achieve a great performance in this domain.

\subsection{The \textsc{Pass} Domain}
\begin{figure}[t]
    \centering
    \includegraphics[width=\linewidth]{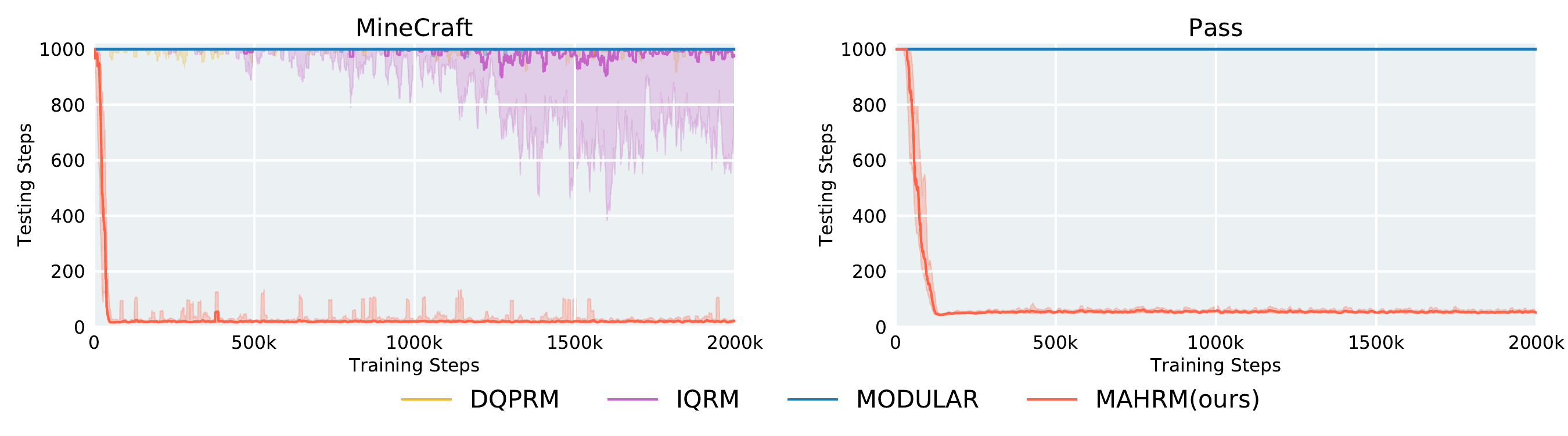}
    \caption{Experimental results in the \textsc{MineCraft} (Left) and the \textsc{Pass} (Right) domains.}
    \label{fig:exp2}
\end{figure}

The last one is the \textsc{Pass} domain mentioned in Subsec. \ref{subsec:HRM}. The prior knowledge utilized in the baselines are the primitive propositions. The RM of the joint task used in DQPRM and IQRM contains 32 states (as illustrated in Appendix A), and thus has a significantly larger size than that of MAHRM. The experimental results in Figure~\ref{fig:exp2} show that MAHRM learns the joint task efficiently while other baselines fail to learn any feasible policies for the agents to pass the room. IQRM still suffers from the issues of non-stationary as the local policy of each agent cannot access the information of other agents. The mechanism of RM projections does not work in DQPRM as the three agents are fully coordinated and highly interdependent. MODULAR cannot choose the correct subtasks to be executed, since the rewards in this domain is sparse and the agents can rarely reach the room through their explorations.

\section{Related Works}
MAHRL approaches have received a rising interest in recent years, due to their potential to decompose long-horizon RL tasks into subtasks with shorter horizon, thus facilitating coordinated learning among multiple agents. \citet{ghavamzadeh2006hierarchical} used a graph with multiple levels to represent the hierarchies of a task, and coordinated the agents by centralized training and execution. 
\citet{tang2018hierarchical} proposed three deep MAHRL architectures, which utilized deep Q-learning, communication network~\cite{sukhbaatar2016learning} and Q-mix network~\cite{rashid2018qmix} to learn a high-level policy for each agent, respectively. Another approach called feudal multi-agent hierarchies proposed by \citet{ahilan2019feudal} is to learn a common central policy for all agents at the highest level. 
\citet{lee2019learning} proposed the modular network consisting of a set of subtasks for each agent and a common meta policy at high-level, and then trained the network using the common team reward and behavior diversification~\cite{eysenbach2018diversity}, without defining the reward functions of subtasks. \citet{yang2019hierarchical} used mutual information to discover diverse subtasks, and \citet{chakravorty2019option} extended the single-agent option-critic framework~\cite{bacon2017option} to multi-agent settings, discovering subtasks by learning termination functions of options using policy gradients. However, these MAHRL approaches do not utilize the prior knowledge of high-level events, and are unable to deal with the issues of non-Markovian rewards.

RMs~\cite{icarte2018using} have been widely used to specify reward functions in RL that cannot be modeled as Markovian ones for more efficient learning. In the fields of robotics, RMs are applied to represent the tasks of vision based robotics~\cite{camacho2020disentangled, camacho2021reward} and quadruped robots~\cite{defazio2021learning}. As a more succinct expression of state-action history based rewards than RMs, temporal logic, especially \textit{Linear Temporal Logic} (LTL)~\cite{pnueli1977temporal} and its extensions, are also incorporated into RL in recent studies~\cite{toro2018teaching,li2017reinforcement,de2018reinforcement,de2020temporal,yuan2019modular,jothimurugan2019composable,leon2020systematic}. \citet{camacho2019ltl} proved that an RM can be constructed from a reward specification representing by LTL formulae that satisfy safe and co-safe properties. \citet{yuan2019modular} leveraged LTL specifications of tasks for modular deep RL, while \citet{leon2020systematic} introduced task temporal logic, an extension of LTL, to help training a deep RL agent via subtasks generalization. \citet{toro2018teaching} proposed an off-policy learning framework to teach a single-agent learn multiple tasks specified by co-safe LTL. However, all these studies only focused on single agent RL problems. \citet{leon2020extended} extended the work~\cite{toro2018teaching} to cooperative MARL settings using independent learning. 
\citet{muniraj2018enforcing} studied zero-sum stochastic games by extending LTL to signal temporal logic, and incorporated it into a deep MARL algorithm based on minimax Q-learning. However, these studies are based on some assumptions in terms of event occurrences and agent independencies, and thus can only work in relatively simple multiagent domains. 

Since hand-crafted RM requires massive human prior knowledge, automatically learning RMs from data is a promising work. \citet{toro2019learning} studied how to learn a perfect RM in partial observable MDP settings by formalizing it as a discrete optimization problem, then solved it via Tabu search. \citet{furelos2020induction} converted agent's trajectories to facts of answer set programming (ASP), then constructed RM from answer set produced by inductive learning of ASP. \citet{gaon2020reinforcement}, \citet{rens2020learning} and \citet{xu2021active} used $L^*$ learning algorithm to learn an RM by assuming that the agent can consecutively answer queries produced from $L^*$. \citet{velasquez2021learning} then made progress with $L^*$ by learning a probabilistic RM where the transitions of RM are non-deterministic. However, to the best of our knowledge, all these works are still limited to single-agent settings, and how to scale up to multi-agent settings is still an open question in the literature.

\section{Conclusions}
In this work, we propose MAHRM for more efficient learning in cooperative MARL problems by decomposing a joint task (\textit{i.e.} propositions) into simple subtasks that are organized as a hierarchical structure. The rewards of these tasks are specified as RMs, and the policies of the subtasks are learned in the way similar to HRL: the policy of a higher-level subtask learns to choose subtasks at lower-level, and the policies at the lowest level determine actions for the agents to execute in the environment.
Experimental results in three domains demonstrate that MAHRM outperforms the MARL baselines that utilize the same prior knowledge of high-level events. While leveraging the predefined RMs is able to facilitate the learning efficiency, it requires massive human prior knowledge. Hence, it is promising to automatically learn the internal structure of RMs from experiences in multi-agent settings, which is left as our future work.

\appendix

\section{Illustrations of RMs}
\begin{figure}[h]
	\centering
	\includegraphics[width=0.5\linewidth]{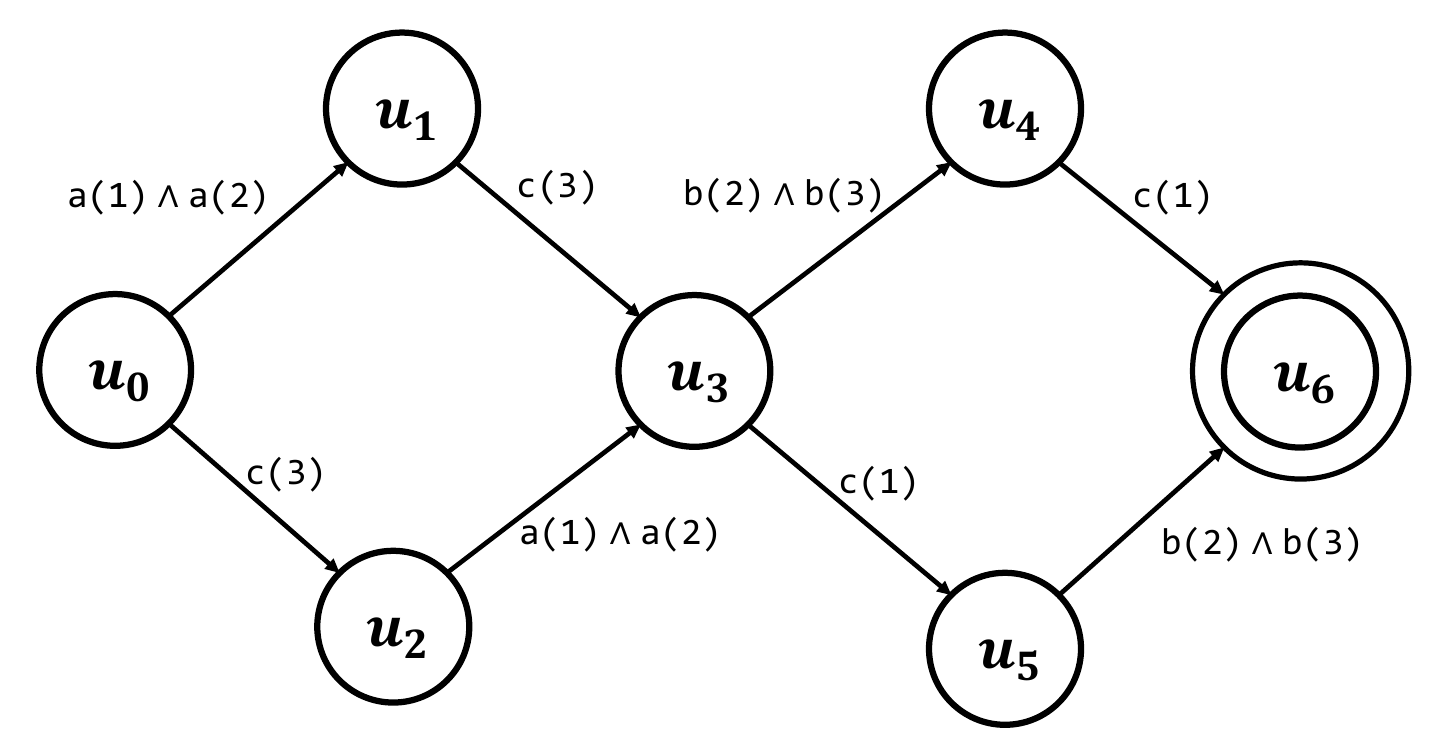}
	\caption{The RM used in the \textsc{MineCraft} domain contains 7 states, with the initial state $u_0$ and the terminal state $u_6$. Its transitions are specified by the primitive propositions \texttt{a(i),b(i)} and \texttt{c(i)}, which means that the objects \texttt{a,b} and \texttt{c} are collected by agent \texttt{i}, respectively (\texttt{i}=1,2,3). It defines the reward function of the joint task described as follows. First, three agents have to complete the following two subtasks in any order: ``\textit{agent 1 and 2 get object} \texttt{a} \textit{simultaneously}'', and ``\textit{agent 3 gets object} \texttt{c}''. After the objects \texttt{a} and \texttt{c} are collected, the RM transits to $u_3$. Then these agents have to complete other two subtasks in any order: ``\textit{agent 2 and 3 get object} \texttt{b} \textit{simultaneously}'', and ``\textit{agent 1 gets object} \texttt{c}''. Finally, the joint task is completed and the RM transits to $u_6$.}
	\label{fig:minecraft_rm}
\end{figure}

\begin{figure}[h]
	\centering
	\includegraphics[width=\linewidth]{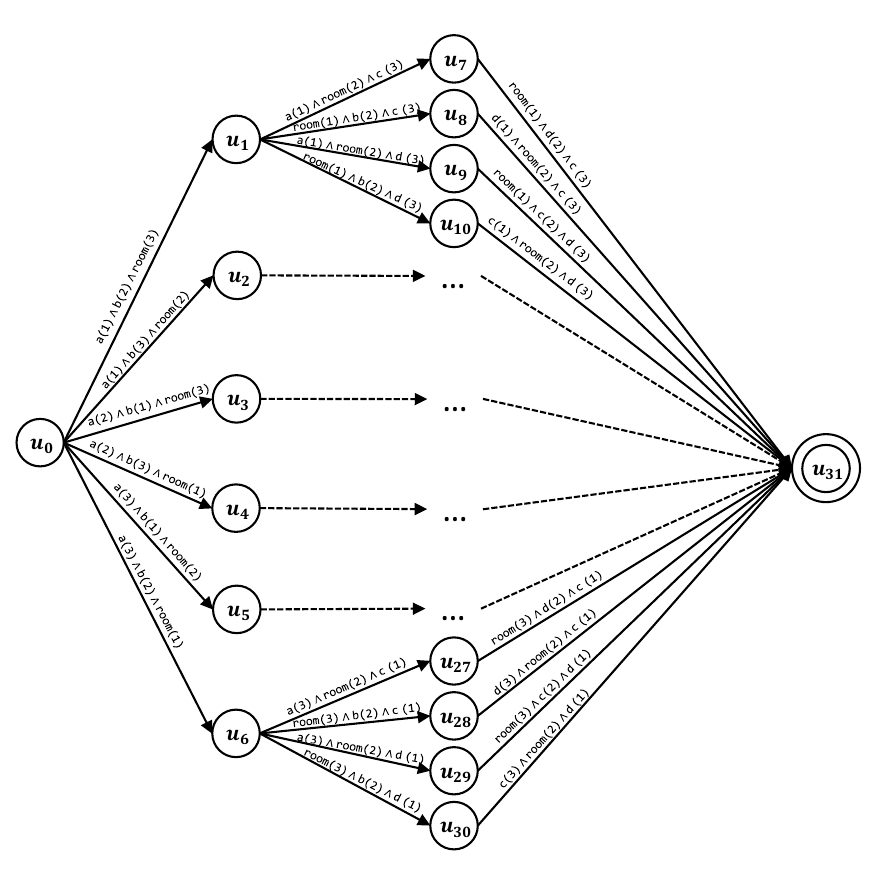}
	\caption{The RM of the joint task used in DQPRM and IQRM in the \textsc{Pass} domain contains 32 states, with the initial state $u_0$ and the terminal state $u_{31}$. The states $u_{11}$-$u_{26}$ are hidden in the figure, and the dotted arrows are their hidden transitions. The transitions of the RM are specified by the primitive propositions \texttt{a(i),b(i),c(i),d(i)} and \texttt{room(i)}, without using the hierarchy of propositions. This RM can be seen as a directed graph with 24 paths from $u_0$ to $u_{31}$, and each path indicates a solution to let three agents pass the door and reach the room. For example, the path $u_0\to u_1\to u_7\to u_{31}$ indicates a solution as follows. First, agent 1 presses button \texttt{a}, agent 2 presses button \texttt{b} and agent 3 reaches the room. Then agent 1 keeps button \texttt{a} pressed, and agent 3 presses button \texttt{c} so as to let agent 2 reach the room. Finally, agent 3 keeps button \texttt{c} pressed, then agent 2 presses button \texttt{d} so as to let agent 1 reach the room.}
	\label{fig:pass_rm}
\end{figure}

The RM of the joint task in the \textsc{MineCraft} domain is demonstrated in Figure \ref{fig:minecraft_rm}. The RM of the joint task used in DQPRM and IQRM in the \textsc{Pass} domain is demonstrated in Figure \ref{fig:pass_rm}. 

\end{document}